\documentclass[a4paper, 10pt, conference]{ieeeconf}      
\usepackage{FG2026}

\FGfinalcopy 

\IEEEoverridecommandlockouts                              
\usepackage{graphicx} 
\usepackage{booktabs} 
\usepackage{amsmath,amssymb}
\usepackage{dblfloatfix}
\usepackage{hyperref}
\def\FGPaperID{232} 

\title{\LARGE \bf
Polyglot: Multilingual Style Preserving Speech-Driven Facial Animation
}

\author{\parbox{16cm}{\centering
    {\large Federico Nocentini$^{1,*}$, Kwanggyoon Seo$^2$, Qingju Liu$^2$, Claudio Ferrari$^1$, Stefano Berretti$^1$, David Ferman$^2$, Hyeongwoo Kim$^3$, Pablo Garrido$^2$, Akin Caliskan$^2$}\\
    {\normalsize
    $^1$ University of Florence, $^2$ Flawless AI, $^3$ Imperial College London}}
    \thanks{$^*$Work done during an internship at Flawless AI}
}

\begin{document}

\ifFGfinal
\thispagestyle{empty}
\pagestyle{empty}
\else
\author{Anonymous FG2026 submission\\ Paper ID \FGPaperID \\}
\pagestyle{plain}
\fi

\maketitle
\begin{figure*}[!t]
  \centering
  \includegraphics[width=0.9\textwidth]{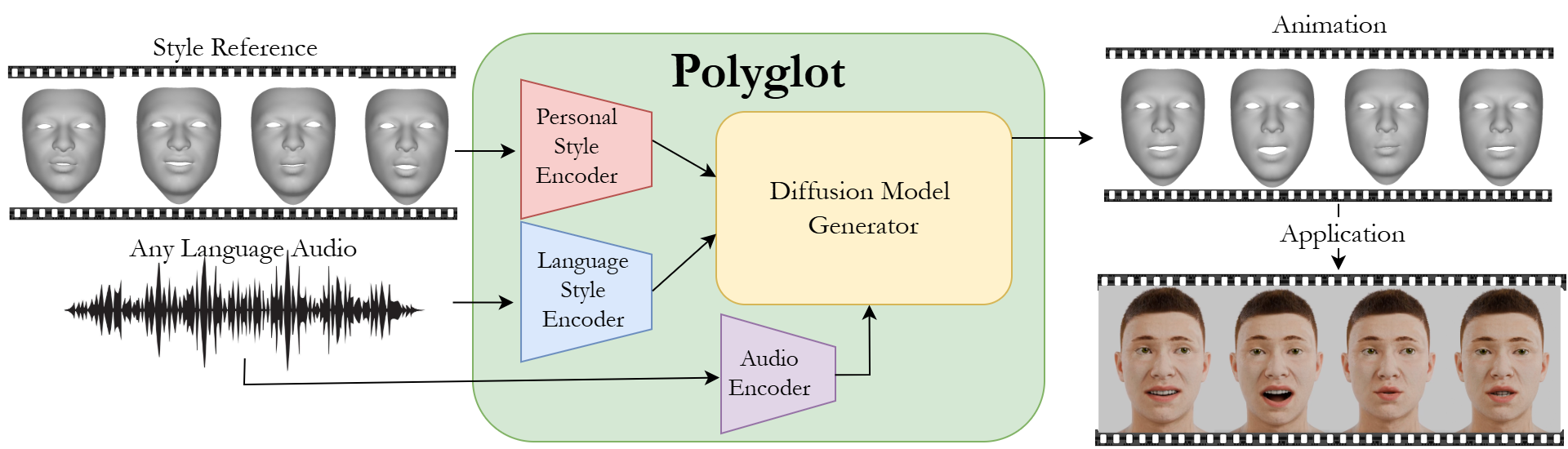}
  \caption{\textbf{Polyglot}, a deep learning architecture for speech-driven facial animation that preserves language and personal speaking styles during animation.}
  \label{fig:teaser}
  \vspace{-0.5cm}
\end{figure*}
\begin{abstract}
Speech-Driven Facial Animation (SDFA) has garnered significant attention in recent years due to its broad range of applications, including movies, video games, and virtual reality. However, off-the-shelf models are trained on single-language audio-animation pairs, limiting their capacity for real-world applications. In this work, we focus on addressing multilingual SDFA that is essential to enhance generation realism, as human speech is deeply influenced by language, shaping not only phonetic properties but also rhythm, intonation, and accompanying facial expressions. Moreover, speaking style is not solely dictated by language. Individuals speaking the same language can exhibit vastly different personal styles. However, simply incorporating multilingual data into training is insufficient. Effective conditioning is crucial, yet most state-of-the-art methods rely on either fixed language-specific labels or fixed/learnable speaker-specific conditioning, not both. This limits their ability to model the nuanced interplay between language and individual speaking styles. To address these limitations, we introduce \textbf{Polyglot}, a single unified diffusion-based architecture designed for personalized multilingual SDFA. Our approach uses speech transcript embeddings to encode language-specific features and leverages speaker-style embeddings, extracted from reference facial sequences, to capture individual speaking habits. Crucially, Polyglot does not require predefined language labels, enabling it to generalize across languages and adapt to intra-language variations. Similarly it does not require any speaker-specific labels, which are learned in a self-supervised learning manner, allowing us to efficiently capture the desired speaking style. By jointly conditioning on both language and speaker style, Polyglot captures subtle expressive traits such as language peculiarities, habitual expressions, rhythm, and articulation patterns. This results in temporally coherent, expressive facial animations that better reflect the speaker’s identity and language characteristics. Our experiments and visual results demonstrate that Polyglot improves animation quality in both monolingual and multilingual scenarios, establishing it as the first approach to effectively unify language and personal style conditioning in SDFA. 
The project website is available at \href{https://fedenoce.github.io/polyglot/}{https://fedenoce.github.io/polyglot/}.
\end{abstract}

\section{Introduction}
\label{sec:intro}
Speech-Driven Facial Animation (SDFA) plays a crucial role in virtual characters, digital assistants, and human-computer interaction, enabling natural and expressive communication through synchronized facial movements and spoken language. While recent advances in deep learning have greatly enhanced the realism of SDFA, most current models operate under monolingual assumptions, limiting their effectiveness in multilingual and multicultural environments~\cite{Fan_Lin_Saito_Wang_Komura_faceformer_2022, nocentini2024scantalk3dtalkingheads, VOCA2019, sun2023diffposetalk, peng2023selftalk, xing2023codetalker}.
However, speech is inherently shaped by language. Phonetics, prosody, rhythm, and intonation vary significantly across languages, influencing both articulation and the broader set of facial expressions accompanying speech~\cite{Ladefoged2000VowelsAC}. For example, tonal languages like Mandarin and Thai need subtle facial movements to help convey word meaning, while stress-timed languages such as English emphasize rhythm and emphasis. Additionally, languages like French and German involve distinct articulation styles, affecting lip, jaw, and facial muscle movements. Beyond linguistic structure, expressive qualities also differ across cultures. Italian and Portuguese often feature melodic intonation and fluid gestures, whereas German and Dutch tend toward more rhythmic and staccato speech patterns.
Moreover, speaking style is not solely dictated by language. Individuals speaking the same language can exhibit vastly different personal styles, including variations in speed, intonation, emotion, and facial expressiveness. While most SDFA methods condition animation solely on audio input, capturing these style-specific features requires incorporating richer conditioning signals.
To address these challenges, we propose \textbf{Polyglot} shown in Fig. \ref{fig:teaser}, the first conditional generative framework based on diffusion models for speech-driven facial animation that jointly models cross-lingual and cross-style variations within a unified architecture. Unlike prior works that consider either language-specific traits~\cite{sungbin2024multitalkenhancing3dtalking} or individual speaking styles~\cite{sun2023diffposetalk}, \textit{Polyglot} integrates both, enabling the generation of facial animations that are linguistically accurate and personally expressive.
In addition, we introduce a new high-quality multilingual dataset consisting of paired audio and sequences of 3D Morphable Model (3DMM) expression parameters, called \textbf{Polyset}. This dataset provides diverse training signals across multiple languages and personal speaking styles, supporting more robust and generalizable learning. Through extensive experiments, we demonstrate that \textit{Polyglot} outperforms state-of-the-art SDFA approaches in multilingual settings, while maintaining strong performance in monolingual scenarios.
Our key contributions are as follows:
\begin{itemize}
    \item We introduce \textit{Polyglot}, a transformer-based diffusion model that generates realistic facial animations by capturing both language and personal speaking styles.
    \item We propose to represent language styles using text embeddings extracted from transcripts of the speech. 
    \item To further encourage preservation of the desired personal speaking style, we introduce a style preservation loss that enhances the retention of speaker-specific characteristics.
\end{itemize}

\section{Related Works}
\label{sec:related}
In recent years, a wide range of models and methods have been developed to tackle the challenge of synchronizing facial animation with speech. Early approaches focused on procedural techniques~\cite{JALI_2016, dominance_Massaro_procedural_2001, Cosi_procedural_2002, Wang_2007_Rulebased_coarticulation_procedural, Xu_2013_diphone_coarticulation_procedural}, relying heavily on visemes (groups of phonemes) to drive facial muscle movements. While effective in controlled scenarios, these systems often require manual retargeting and struggle to generalize without significant preprocessing.

With the rise of large datasets and more powerful models, statistical and deep learning approaches have taken center stage~\cite{VOCA2019, richard2021meshtalk, Fan_Lin_Saito_Wang_Komura_faceformer_2022, facexhubert, Thambiraja_2023_ICCV_imitator, thambiraja2023_3diface, landmarks_3D_Nocentini_2023, xing2023codetalker, FaceDiffuser_Stan_MIG2023, nocentini2024scantalk3dtalkingheads, nocentini2024emovocaspeechdrivenemotional3d, nocentini2024fixedtopologiesunregisteredtraining, nocentini2026freetalkemotionaltopologyfree3d}. VOCA~\cite{VOCA2019} marked an early breakthrough by training on high-quality audio-mesh pairs, but was soon surpassed by MeshTalk~\cite{richard2021meshtalk}, which used the larger, registered Multiface dataset~\cite{wuu2022multiface} and introduced a more expressive model.
Subsequent models further advanced the field through architectural innovations and improved audio encoders. FaceFormer~\cite{Fan_Lin_Saito_Wang_Komura_faceformer_2022} applied a Transformer architecture with an autoregressive design and used the pretrained audio encoder Wav2Vec2~\cite{wav2vec_2019}, a strategy later extended by CodeTalker~\cite{xing2023codetalker}. More recently, FaceXHubert~\cite{facexhubert} and FaceDiffuser~\cite{FaceDiffuser_Stan_MIG2023} adopted HuBERT~\cite{Hubert_audio_encoding}, an enhanced audio encoder that improved the mapping from audio to facial motion.
These innovations have also opened doors for modeling additional factors such as emotion~\cite{peng2023emotalk, danvevcek2023emotional, Wu_2024} and head pose~\cite{SadTalker_2023, sun2023diffposetalk}. However, the scarcity of high-quality, registered 3D data remains a core challenge. To address data limitations, SelfTalk~\cite{peng2023selftalk} proposed a self-supervised framework combining lip-reading and text-to-audio generation. Similarly, KMTalk~\cite{xu2024kmtalkspeechdriven3dfacial} introduced a coarse-to-fine design that combines linguistic priors with learned representations.
Recent work has also tackled issues related to facial topology and representation. ScanTalk~\cite{nocentini2024scantalk3dtalkingheads} proposed an encoder-decoder that supports arbitrary 3D face topologies, while UniTalker~\cite{fan2024unitalkerscalingaudiodriven3d} expanded on this by developing a unified representation framework.

Diffusion models have recently gained traction in SDFA. FaceDiffuser~\cite{FaceDiffuser_Stan_MIG2023} pioneered their use for facial animation, and DiffPoseTalk~\cite{sun2023diffposetalk} extended this with style injection and head pose prediction, demonstrating the potential of diffusion-based generation.
In multilingual contexts, MultiTalk~\cite{sungbin2024multitalkenhancing3dtalking} is, to our knowledge, the only prior work tackling SDFA across multiple languages within a single model. Their approach used 2D video data with pseudo-3D mesh reconstructions and represented language through a learnable lookup table with fixed dictionary embeddings, inspired by CodeTalker.
In contrast, our proposed method, \textit{Polyglot}, is the first to combine multilingual speech-driven facial animation with diffusion models. Instead of relying on fixed language embeddings, we condition generation on rich text embeddings derived from transcripts, allowing the model to capture nuanced linguistic structures. This helps Polyglot better distinguish between linguistically distant languages (e.g., Mandarin vs. English), while leveraging shared features in closely related ones (e.g., Spanish vs. Catalan).
In addition, our model conditions generation on a style embedding that captures personal speaking traits. Unlike earlier models such as VOCA~\cite{VOCA2019}, FaceFormer~\cite{Fan_Lin_Saito_Wang_Komura_faceformer_2022}, or CodeTalker~\cite{xing2023codetalker}, which use one-hot speaker encodings, we follow recent advances~\cite{sun2023diffposetalk, zhang2024personatalkbringattentionpersona, guan2024resyncerrewiringstylebasedgenerator, ao2023gesturediffuclipgesturediffusionmodel} in using encoder-based style embeddings. This approach enhances generalization to unseen speakers and enables a richer, more personalized style representation.
Overall, we show that conditioning on contextual text and speaker style embeddings leads to more expressive, language-aware facial animations. Our approach not only improves performance across languages, but also enables more controllable and diverse animation generation at inference time.

\begin{figure*}[h]
    \centering
    \includegraphics[width=\textwidth]{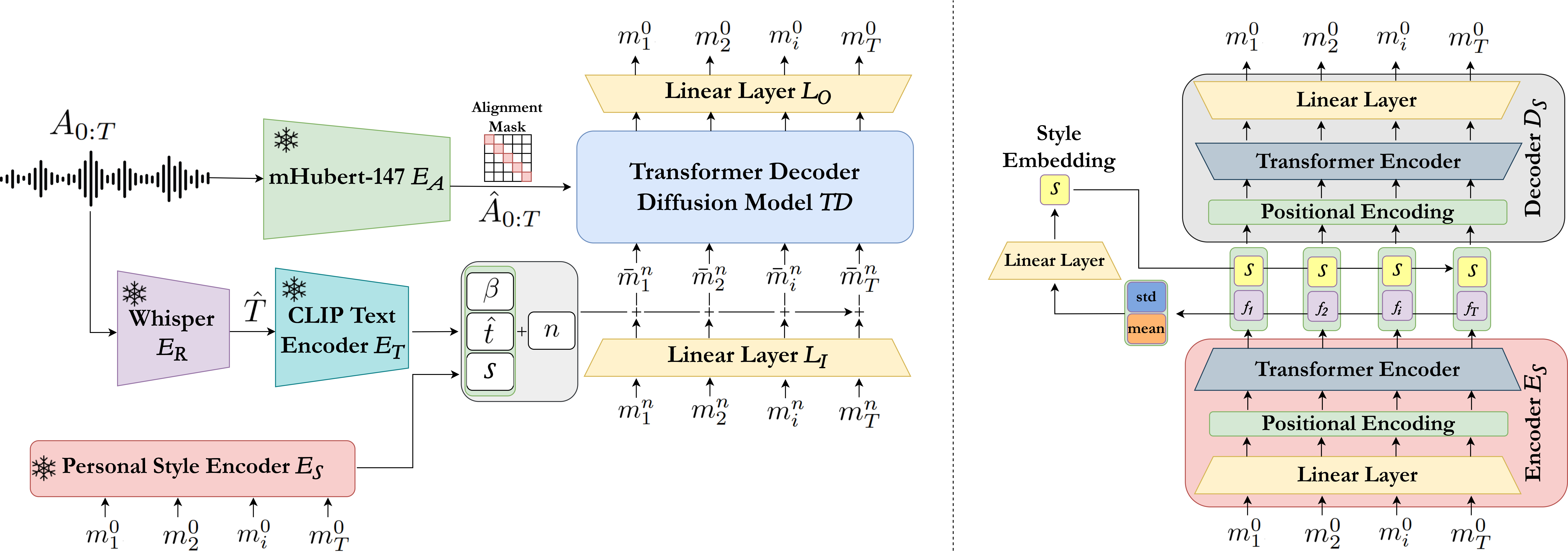}
    \caption{\textit{Left}: \textbf{Polyglot} architecture. Audio \( A_{0:T} \) is processed by mHuBERT \( E_A \), Whisper \( E_R \), and CLIP \( E_T \) to extract features, transcripts, and language embeddings. A style embedding \( S \) is computed from input motion \( M_{0:T}^0 \) via style encoder \( E_S \). Conditioned on identity \( \beta \), language \( \hat{t} \), style \( S \), and timestep \( n \), the diffusion decoder \( TD \) denoises noisy parameters \( M^n \) into motion \( M^0 \).  
    \textit{Right}: The style encoder \( E_S \) extracts per-frame features and pools them into \( S \). The decoder \( D_S \) reconstructs motion \( M_{0:T}^0 \) conditioned on \( S \) to preserve style. This module is trained separately; \( E_S \) is frozen during Polyglot training.
    }

    \vspace{-0.5cm}
    \label{fig:architecture}
\end{figure*}

\section{Proposed Method}
\label{sec:method}
We present \textit{\textbf{Polyglot}}, a multilingual, style-preserving speech-driven facial animation framework. It uses a pre-trained multilingual speech encoder to extract expressive audio features and represents facial motions with 3D Morphable Model (3DMM) parameters~\cite{blanz1999morphable}. A transformer-based denoising network models temporal dynamics via reverse diffusion, guided by classifier-free conditioning. Language is encoded through transcript-based text embeddings, and speaker style via a pre-trained encoder. By jointly modeling language and personal style, \textit{Polyglot} produces expressive, linguistically accurate animations across languages and speakers.

\paragraph{General idea} \textit{Polyglot} is an encoder-decoder that generates 3DMM expression parameters from speech, shape, and a reference sequence capturing speaking style (see Fig. ~\ref{fig:architecture}). It leverages four pretrained modules: a multilingual speech encoder, an Automatic Speech Recognition (ASR) model, a text encoder, and a style encoder, providing rich conditioning. During training, the model learns to denoise expression sequences; at inference, it generates animations by refining noise with these conditions. ICT 3DMM~\cite{ict} is used for face representation; details are in the supplementary.

\paragraph{Personal Style Autoencoder} 
To extract a compact and expressive style embedding, we employ an autoencoder composed of an encoder \(E_S\) and a decoder \(D_S\), with their architectures shown in Fig. \ref{fig:architecture}. During training the encoder takes as input the ground-truth expression parameter sequence \(M_{1:T}^0 = (m_1^0, m_2^0, \dots, m_T^0) \in \mathbb{R}^{T \times k}\), where \(k\) denotes the number of 3DMM expression parameters per-frame, and outputs temporal features \(f_{1:T} = E_S(M_{1:T}^0) \in \mathbb{R}^{T \times h} \), where \( h \) is the latent dimensionality of the autoencoder. 
The global style embedding \(S  \in \mathbb{R}^{h}\) is obtained by concatenating and linearly interpolating the mean and standard deviation of \(f_{1:T}\), following the approach in~\cite{zhang2024personatalkbringattentionpersona}. The decoder learns to reconstructs the expression sequence as \(\hat{M}_{1:T}^0 = D_S(f_{1:T} \oplus S)\), where \(\oplus\) denotes concatenation. 
The autoencoder is trained using a mean squared error loss defined as \(L = \left\| M_{1:T}^0 - \hat{M}_{1:T}^0 \right\|_2^2\). Once trained, the encoder \(E_S\) is used to extract individual style embeddings to condition the generation process in Polyglot.

\paragraph{Language Style Encoder}  
To capture the linguistic structure and style of spoken language, we use a combination of transcription and text encoding. The audio signal is first transcribed into text using the pretrained ASR model \(E_R\) Whisper~\cite{radford2021learningtransferablevisualmodels}, which is chosen for its robustness in multilingual contexts. The transcript is then passed to a pretrained 
CLIP~\cite{clip} text encoder \(E_T\), which produces a sentence-level embedding.  
Despite being trained primarily on English, the CLIP encoder provides rich semantic representations that generalize well across languages. The resulting text embedding captures language-aware nuances that help Polyglot model phonetic, rhythmic, and prosodic features across different languages. This embedding serves as one of the key conditioning inputs to the diffusion-based generation model. More details in ~\ref{sec:quali}.

\paragraph{Problem Formulation}  
Let our multilingual training set consist of samples where \( A_{1:T} \) is an audio sample of a spoken sentence over a temporal window of \( T \) frames, \( \beta \) is the corresponding identity shape parameter of a template face, and \( M_{1:T}^{0} = (m_1^0, m_2^0, \dots, m_T^0) \in \mathbb{R}^{T \times k} \) represents the ground-truth expression parameters sequence, with \( k \) denoting the number of 3DMM expression parameters per frame. Our goal is to learn a mapping function that predicts the expression sequence from the audio, identity parameter \( \beta \), and personal speaking style \( S \), formally expressed as:
\begin{equation}
\text{Polyglot}(A_{1:T}, \beta, S) \approx M_{1:T}^{0}.
\end{equation}
To model this mapping , we rely on four pretrained submodules. We use mHubert-147~\cite{boito2024mhubert} for \( E_A \), due to its effectiveness in multilingual settings and its efficiency in both training and inference. For transcription, we select Whisper~\cite{radford2021learningtransferablevisualmodels}, which is known for its robustness even in non-Latin scripts. Despite being trained only on English, the CLIP text encoder~\cite{clip} used as \( E_T \) provides rich sentence-level embeddings that help model language-aware generation. The personal speaking style encoder \( E_S \) is trained separately in an autoencoder fashion to reconstruct expression sequences from themselves, effectively learning a speaker-specific embedding from the input.
Given an input audio clip \( A_{0:T} \), we extract per-frame features \( \hat{A}_{1:T} = E_A(A_{1:T}) \) to serve as temporal guidance. Simultaneously, the ASR model produces a transcript \( \hat{T} = E_R(A_{1:T}) \), which is transformed by the text encoder into a transcript embedding \( \hat{t} = E_T(\hat{T}) \). To encode a speaking style, the reference expression sequence is passed through \( E_S \), yielding a style embedding \( S = E_S(M_{1:T}^{0}) \).
We generate speech-driven facial animations using a diffusion model that consists of two stages: a forward process that gradually adds Gaussian noise to the clean data \( M_{1:T}^{0} \), and a reverse process that learns to reconstruct the clean expression sequence from the noisy input. Formally, the forward process is modeled as a Markov chain \( q(M_{1:T}^{n} | M_{1:T}^{n-1}) \), where \( n \in \{1, \dots, N\} \) denotes the diffusion step. The reverse process, which approximates the intractable posterior \( q(M_{1:T}^{n-1} | M_{1:T}^{n}) \), is learned by a denoising network trained to predict \( M_{1:T}^{0} \) given \( M_{1:T}^{n} \).
To implement the reverse process, we adopt a transformer decoder (\textit{TD}) similar to~\cite{sun2023diffposetalk}, which includes positional encoding, multi-head self-attention, and multi-head cross-attention mechanisms. As in prior work~\cite{Fan_Lin_Saito_Wang_Komura_faceformer_2022, Thambiraja_2023_ICCV_imitator, xing2023codetalker}, we use an alignment mask to guide the cross-attention between the inputs and the audio features.
Our model extends~\cite{sun2023diffposetalk} by introducing a latent conditioning mechanism at each denoising step. Specifically, a linear layer \( L_I \) maps the noisy expression sequence \( M_{1:T}^n \) into the transformer decoder’s latent space. The input to the decoder is then given by:

\begin{equation}
\bar{M}_{1:T}^n = L_I(M_{1:T}^n) + c, \quad \text{where} \quad c = L_1(\hat{t} \oplus \beta \oplus S) + L_2(n).
\end{equation}
Here, \( \oplus \) denotes the concatenation operation, and \( L_1 \), \( L_2 \) are linear projections. \( L_1 \) maps the concatenated text embedding \( \hat{t} \), identity parameter \( \beta \), and personal speaking style embedding \( S \) into the decoder’s hidden space, while \( L_2 \) encodes the diffusion step \( n \). The final conditioned input to the decoder is \( \bar{M}_{1:T}^n = (\bar{m}_1^n, \bar{m}_2^n, \dots, \bar{m}_T^n) \in \mathbb{R}^{T \times h} \), where \( h \) is the latent dimensionality of the decoder.
This formulation enables our model to generate expressive and language-aware 3D facial animations while preserving personal speaking styles across multiple languages.
\paragraph{Training Strategy} 
Following the method in \cite{sun2023diffposetalk}, we partition a speech feature sequence of length \( T \) into windows of size \( T_w \). To ensure smooth transitions between consecutive windows, we use the last \( T_p \) frames of speech features \( \hat{A}_{-T_p:0} \) and the conditioned expression parameters sequence \( \bar{M}_{-T_p:1}^0 \) from the preceding window as conditional inputs. For the first window, we replace the speech features and conditioned expression parameters sequences with learnable start features \( A_{\text{start}} \) and \( M_{\text{start}} \). At each diffusion step \( n \), the network receives both the previous and current speech features \( \hat{A}_{-T_p:T_w} \), the previous conditioned expression parameters sequence \( \bar{M}_{-T_p:1}^0 \), and the current noisy expression parameters sequence \( \bar{M}_{1:T_w}^n \), sampled from \( q(M_{1:T_w}^n | M_{1:T_w}^0) \). The denoising network then produces the clean sample. Similar to the input, we apply a linear layer \( L_O \) to map the model's prediction from the decoder’s latent space into the expression parameters sequence space (Fig. \ref{fig:architecture}):
\begin{equation}
\hat{M}_{-T_p:T_w}^0 = L_O\left(TD(\bar{M}_{-T_p:1}^0, \bar{M}_{1:T_w}^n,  \hat{A}_{-T_p:1},  \hat{A}_{1:T_w})\right).
\end{equation}
\noindent
To train the model, inspired by previous works~\cite{sun2023diffposetalk, nocentini2024fixedtopologiesunregisteredtraining, Thambiraja_2023_ICCV_imitator, Fan_Lin_Saito_Wang_Komura_faceformer_2022, xing2023codetalker}, we employ a combination of losses in both the expression parameters sequence space and the vertex space. 
To guide the model in learning accurate and expressive facial dynamics, we define a unified loss function \( L_{\text{simple}} \) over the expression parameters sequence. This loss supervises the overall expression prediction while placing additional emphasis on the subset of parameters responsible for mouth movements. It is defined as:
\begin{equation}
\begin{split}
L_{\text{simple}} &= \left\| \hat{M}_{-T_p:T_w}^0 - M_{-T_p:T_w}^0 \right\|_2^2 + \\
&\quad + \lambda_{\text{mouth}} \left\| \hat{B}_{-T_p:T_w}^0 - B_{-T_p:T_w}^0 \right\|_2^2
\end{split}
\end{equation}
where \( \hat{M}_{-T_p:T_w}^0 \) and \( M_{-T_p:T_w}^0 \) denote the predicted and ground-truth expression parameter sequences, respectively, and \( \hat{B}_{-T_p:T_w}^0 \), \( B_{-T_p:T_w}^0 \) correspond to the predicted and ground-truth mouth-related blendshapes. The hyperparameter \( \lambda_{\text{mouth}} \) controls the relative importance of mouth accuracy in the loss.
To further encourage the model to preserve the personal speaking style, we introduce a style preservation loss. This loss minimizes the difference between the style embedding provided as input and the embedding inferred from the generated expression parameters sequence:
\begin{equation}
L_{\text{style}} = \left\| E_S(\hat{M}_{-T_p:T_w}^0) - E_S(M_{1:T}^0) \right\|_2^2,
\end{equation}
Additionally, to better constrain the facial motion in 3D space, we convert both the predicted and ground-truth expression parameters into 3D mesh sequences, denoted by \( \hat{V}_{-T_p:T_w}^0 \) and \( V_{-T_p:T_w}^0 \), respectively. We then apply geometric losses, including a vertex loss \( L_{\text{vert}} \) that enforces spatial accuracy at the vertex level, a velocity loss \( L_{\text{vel}} \) that ensures temporal coherence between consecutive frames, and a smoothness loss \( L_{\text{smooth}} \) that penalizes large accelerations to encourage natural and stable motion.
They are defined as follows:
\begin{gather}
L_{\text{v}} = \left\| \hat{V}_{-T_p:T_w}^0 - V_{-T_p:T_w}^0 \right\|_2^2, \\
\hat{D} = \hat{V}_{-T_p+1:T_w}^0 - \hat{V}_{-T_p:T_w-1}^0, \\
D = V_{-T_p+1:T_w}^0 - V_{-T_p:T_w-1}^0, \\
L_{\text{vel}} = \left\| \hat{D} - D \right\|_2^2, \\
L_{\text{smooth}} = \left\| \hat{V}_{-T_p+2:T_w}^0 - 2\hat{V}_{-T_p+1:T_w-1}^0 + \hat{V}_{-T_p:T_w-2}^0 \right\|_2^2.
\end{gather}
Finally, our total loss function is expressed as:
\begin{equation}
L =  \lambda_{\text{sim}}L_{\text{simple}} + \lambda_{\text{style}} L_{\text{style}} + \lambda_{\text{v}} L_{\text{v}} + \lambda_{\text{vel}} L_{\text{vel}} + \lambda_{\text{s}} L_{\text{smooth}} .
\end{equation}
\noindent
\paragraph{Inference Strategy} During the synthesis process, we iteratively generate the output sequence \( \hat{M}_{1:T}^0 \) based on \( (A_{0:T}, \hat{t}, \beta, S) \). Specifically, we approximate the clean sample using the function \(\hat{M}_{1:T}^0 = Polyglot(\hat{M}_{1:T}^n, A_{1:T}, \beta, \hat{t}, S, n)\), then reintroduce noise to derive \( \hat{M}_{1:T}^{n-1} \). This process is repeated for \( n = N, N-1, \dots, 1 \), if \(n=N\), then \(\hat{M}_{1:T}^N \sim \mathcal{N}(0, 1)\). Thus, during the sampling phase, our method begins with audio features, identity shape parameters, individual style embedding and Gaussian noise to generate the desired animation. Furthermore, we incorporate classifier-free guidance~\cite{ho2022classifierfreediffusionguidance} in conjunction with the incremental framework~\cite{brooks2023instructpix2pixlearningfollowimage},
which has proven effective in generating images influenced by multiple conditioning signals and has also demonstrated success in audio-driven facial animation~\cite{sun2023diffposetalk}. 

\section{Experimental Results}
\label{sec:experiments}
In the following, we first introduce the datasets and evaluation metrics in Sec.~\ref{sec:dataset} and Sec.~\ref{sec:metrics}, respectively. We then present quantitative and qualitative results, comparing our approach with state-of-the-art methods in Sec.~\ref{sec:quanti} and Sec.~\ref{sec:quali}. In Sec.~\ref{sec:abl}, we conduct ablation studies to analyze the impact of different architectural components. Finally, we report the results of a user study comparing our method with state-of-the-art approaches in Sec.~\ref{sec:us}.

\begin{table*}[h!]
  \centering
  \caption{Polyglot in comparison with SOTA methods. All models listed in the table were trained on a multilingual dataset spanning 20 languages. We then evaluated each model in five monolingual settings and a multilingual settings (20 Languages) to assess their performance across different linguistic conditions (best values in bold, second best underlined).
        }
    \resizebox{\linewidth}{!}{
    \centering
    \begin{tabular}{@{}l@{\hspace{0.1cm}}ccccccccccccccc@{}} 
        \toprule
        & \multicolumn{4}{c}{\textbf{Italian}} & \phantom{abc} & \multicolumn{4}{c}{\textbf{Spanish}} &
        \phantom{abc} & \multicolumn{4}{c}{\textbf{English}} 
        \\
        \cmidrule{2-5} \cmidrule{7-10} \cmidrule{12-15}
         & LVE $\downarrow$ & MVE $\downarrow$ & DTW $\downarrow$ & MOD $\downarrow$ && LVE $\downarrow$ & MVE $\downarrow$ & DTW $\downarrow$ & MOD $\downarrow$ && LVE $\downarrow$ & MVE $\downarrow$ & DTW $\downarrow$ & MOD $\downarrow$\\
        \midrule
        FaceFormer~\cite{Fan_Lin_Saito_Wang_Komura_faceformer_2022} & 0.348 & 0.061 & 0.171 & 0.266 && 0.436 & 0.081 & 0.189 & 0.299 && 0.425 & 0.082 & 0.191 & 0.313 \\
        
        SelfTalk~\cite{peng2023selftalk} & 0.475 & 0.098 & 0.202 & 0.307 && 0.593 & 0.119 & 0.225 & 0.335 && 0.629 & 0.126 & 0.230 & 0.367 \\

        DiffPoseTalk~\cite{sun2023diffposetalk} & 0.290 & 0.065 & 0.172 & 0.253 && 0.419 & 0.084 & 0.191 & 0.314 && \underline{0.406} & 0.083 & 0.194 & 0.305\\
        
        MultiTalk~\cite{sungbin2024multitalkenhancing3dtalking} & 0.312 & 0.071 & 0.168 & 0.232 && 0.420 & 0.075 & 0.181 & 0.276 && 0.439 & 0.113 & 0.199 & 0.319 \\

        \midrule

        S-Faceformer~\cite{guan2024resyncerrewiringstylebasedgenerator} & 0.297 & \underline{0.058} & 0.161 & 0.242 && 0.340 & 0.069 & 0.173 & 0.292 && \underline{0.333} & \underline{0.066} & \underline{0.167} & 0.294 \\

        S-DiffPoseTalk~\cite{sun2023diffposetalk} & 0.293 & 0.060 & \underline{0.160} & 0.237 && \underline{0.330} & \underline{0.068} & \underline{0.168} & \underline{0.266} && 0.346 & 0.070 & 0.173 & \underline{0.272} \\
        
        \midrule

        Polyglot & \textbf{0.237} & \textbf{0.049} & \textbf{0.141} & \textbf{0.202} && \textbf{0.306} & \textbf{0.063} & \textbf{0.152} & \textbf{0.248} && \textbf{0.292} & \textbf{0.059} & \textbf{0.155} & \textbf{0.256} \\
        
        \bottomrule

        & \multicolumn{4}{c}{\textbf{Portuguese}} & \phantom{abc} & \multicolumn{4}{c}{\textbf{Japanese}} &
        \phantom{abc} & \multicolumn{4}{c}{\textbf{20 Languages}} 
        \\
        \cmidrule{2-5} \cmidrule{7-10} \cmidrule{12-15} 
         & LVE $\downarrow$ & MVE $\downarrow$ & DTW $\downarrow$ & MOD $\downarrow$ && LVE $\downarrow$ & MVE $\downarrow$ & DTW $\downarrow$ & MOD $\downarrow$ && LVE $\downarrow$ & MVE $\downarrow$ & DTW $\downarrow$ & MOD $\downarrow$\\
        \midrule
        FaceFormer~\cite{Fan_Lin_Saito_Wang_Komura_faceformer_2022} & 0.325 & 0.064 & 0.162 & 0.272 && 0.222 & 0.044 & 0.145 & 0.231 && 0.290 & 0.057 & 0.158 & 0.243 \\
        
        SelfTalk~\cite{peng2023selftalk} & 0.492 & 0.103 & 0.200 & 0.317 && 0.293 & 0.060 & 0.162 & 0.243 && 0.405 & 0.082 & 0.183 & 0.264 \\

        DiffPoseTalk~\cite{sun2023diffposetalk} & 0.337 & 0.072 & 0.176 & 0.263 && 0.223 & 0.044 & 0.146 & 0.232 && 
        0.275 & 0.056 & 0.158 & 0.227 \\
        
        MultiTalk~\cite{sungbin2024multitalkenhancing3dtalking} & 0.317 & 0.069 & 0.184 & 0.296 && 0.237 & 0.051 & 0.156 & 0.241 && 0.286 & 0.060 & 0.161 & 0.231  \\
        
        \midrule

        S-Faceformer~\cite{guan2024resyncerrewiringstylebasedgenerator} & \underline{0.273} & \underline{0.056} & 0.154 & 0.241 && 0.200 & 0.039 & 0.137 & 0.247 && 0.249 & 0.049 & 0.149 & 0.232 \\

        S-DiffPoseTalk~\cite{sun2023diffposetalk} & 0.275 & 0.057 & \underline{0.150} & \underline{0.235} && \underline{0.166} & \underline{0.034} & \underline{0.123} & \underline{0.191} && \underline{0.235} & \underline{0.048} & \underline{0.142} & 
        \underline{0.205} \\
        
        \midrule

        Polyglot & \textbf{0.239} & \textbf{0.050} & \textbf{0.131} & \textbf{0.216} && \textbf{0.137} & \textbf{0.028} & \textbf{0.111} & \textbf{0.171} && \textbf{0.199} & \textbf{0.041} & \textbf{0.126} & \textbf{0.188} \\

        \bottomrule
        
        \end{tabular}
        }
        \label{tab:quantitative}
\end{table*}

\subsection{PolySet Dataset}
\label{sec:dataset}
We present \textbf{PolySet}, a new dataset derived from the MultiTalk~\cite{sungbin2024multitalkenhancing3dtalking} 2D video corpus.  We plan to release PolySet, which includes paired sequences of audio and expression parameters. MultiTalk comprises 420 hours of talking head videos spanning 20 languages. While MultiTalk was originally collected from YouTube and curated for frontal face verification and active speaker identification, we further refine the dataset by removing samples with noisy audio.
Our further cleaning process aims to eliminate samples that could negatively impact model training and results. To ensure high-quality data, we applied filtering based on two criteria: \textit{PESQ}~\cite{pesq,kumar2023torchaudio} metric for assessing audio quality, and reprojected landmark error (\textit{RLE}) for the accuracy of generated expression parameters. We removed all the samples with \textit{PESQ} lower than 2, and \textit{RLE} larger than 5 pixels.
For each frame, we extract 3DMM expression and shape parameters using the ICT topology~\cite{ict}. 
After filtering, we obtain a balanced dataset of 10,000 sentences across 20 languages, at 25fps, with each language containing 450 sentences for training and 50 for validation and testing, resulting in a total of 16 hours of clean data. 
\subsection{Evaluation Metrics}
\label{sec:metrics}

We evaluate Polyglot against state-of-the-art approaches using four key metrics widely adopted in prior works~\cite{facexhubert, FaceDiffuser_Stan_MIG2023, Fan_Lin_Saito_Wang_Komura_faceformer_2022, nocentini2024scantalk3dtalkingheads, nocentini2024emovocaspeechdrivenemotional3d, nocentini2024fixedtopologiesunregisteredtraining, Thambiraja_2023_ICCV_imitator, sun2023diffposetalk}. Although our model predicts expression parameters, evaluations are conducted in the vertex space for ease of interpretation and comparison with previous works:  

\begin{itemize}
    \item \textbf{LVE} and \textbf{MVE}~\cite{Fan_Lin_Saito_Wang_Komura_faceformer_2022, facexhubert, landmarks_3D_Nocentini_2023}: Measure max and mean $\ell_2$ lip vertex error, respectively, quantifying per-frame deviations from the ground-truth meshes.  
    \item \textbf{DTW}~\cite{nocentini2024fixedtopologiesunregisteredtraining, Thambiraja_2023_ICCV_imitator}: Computes lip-sync distance using Dynamic Time Warping, assessing temporal alignment with speech.  
    \item \textbf{MOD}~\cite{sun2023diffposetalk}: Evaluates mouth opening discrepancies to measure articulation expressiveness.  
\end{itemize}

\subsection{Quantitative Results}\label{sec:quanti}
In Table~\ref{tab:quantitative}, we report a quantitative comparison between Polyglot and several state-of-the-art methods, all trained on the PolySet dataset. We include comparisons with models that incorporate language information, such as MultiTalk~\cite{sungbin2024multitalkenhancing3dtalking}, as well as those that utilize personal speaking style information, including S-Faceformer~\cite{zhang2024personatalkbringattentionpersona} and S-DiffPoseTalk~\cite{sun2023diffposetalk}. Each column in the table corresponds to a different test set, covering both single-language and multilingual scenarios.
Polyglot, DiffPoseTalk, and S-DiffPoseTalk~\cite{sun2023diffposetalk} are trained directly on 3DMM expression parameters, while the remaining methods are vertex-based and rely on 3D face mesh sequences derived from these parameters. The results demonstrate that Polyglot, which is explicitly designed for multilingual settings, consistently outperforms existing approaches in both single-language and multilingual evaluations. This underscores the benefits of incorporating spoken language information into the architecture, as it significantly enhances performance.
Furthermore, models that do not explicitly account for multilingual input tend to underperform in multilingual settings compared to architectures that are designed to process audio across different languages. As shown in Table~\ref{tab:quantitative}, methods that leverage personal style embeddings, including Polyglot, achieve superior performance compared to those that rely solely on audio input. Notably, Polyglot surpasses models based on language features, such as MultiTalk~\cite{sungbin2024multitalkenhancing3dtalking}, as well as models focused on personal speaking style, such as S-Faceformer~\cite{zhang2024personatalkbringattentionpersona} and S-DiffPoseTalk~\cite{sun2023diffposetalk}. A detailed comparison for every language is provided in the supplementary. 

\subsection{Qualitative Results}
\label{sec:quali}

\begin{figure*}[!ht]
    \centering
    \begin{minipage}[b]{0.52\textwidth}
        \centering
        \includegraphics[width=\linewidth]{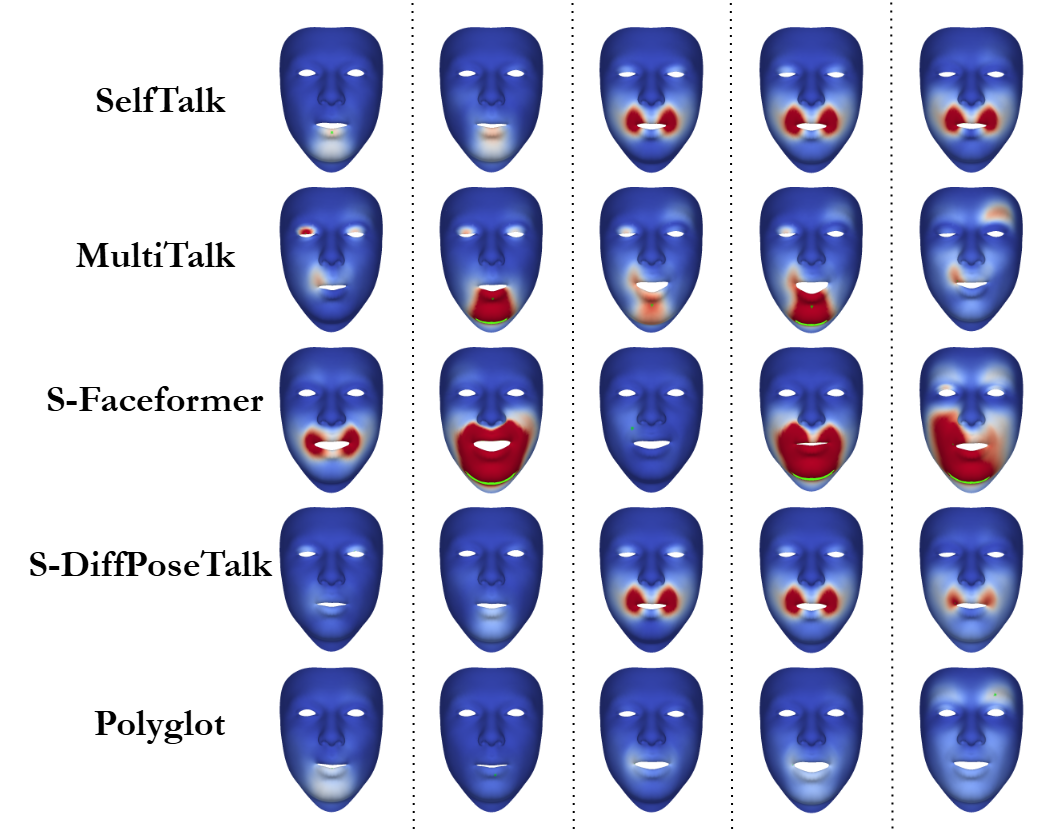}
        \caption{Japanese.}
        \label{fig:heatmap_1}
    \end{minipage}
    \hfill
    \vline 
    \hfill
    \begin{minipage}[b]{0.44\textwidth}
        \centering
        \includegraphics[width=\linewidth]{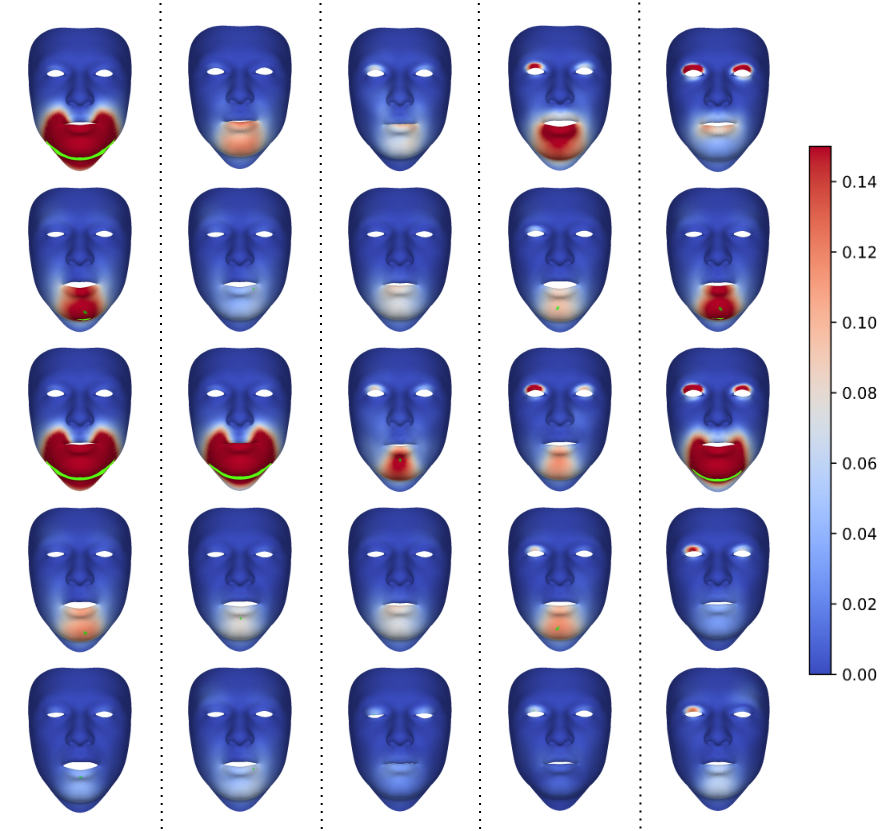}
        \caption{Polish.}
        \label{fig:heatmap_2}
    \end{minipage}
    
    \caption{Heatmap visualization of the \( \ell_2 \) vertex error between predictions and ground truth. We compare Polyglot with state-of-the-art methods, where blue regions indicate lower errors and red regions indicate higher errors. Results are shown for two languages: (a) Japanese and (b) Polish.}
    \label{fig:heatmaps}
    \vspace{-0.5cm}
\end{figure*}

\begin{figure}[!ht]
    \centering
    \centering
    \includegraphics[width=\linewidth]{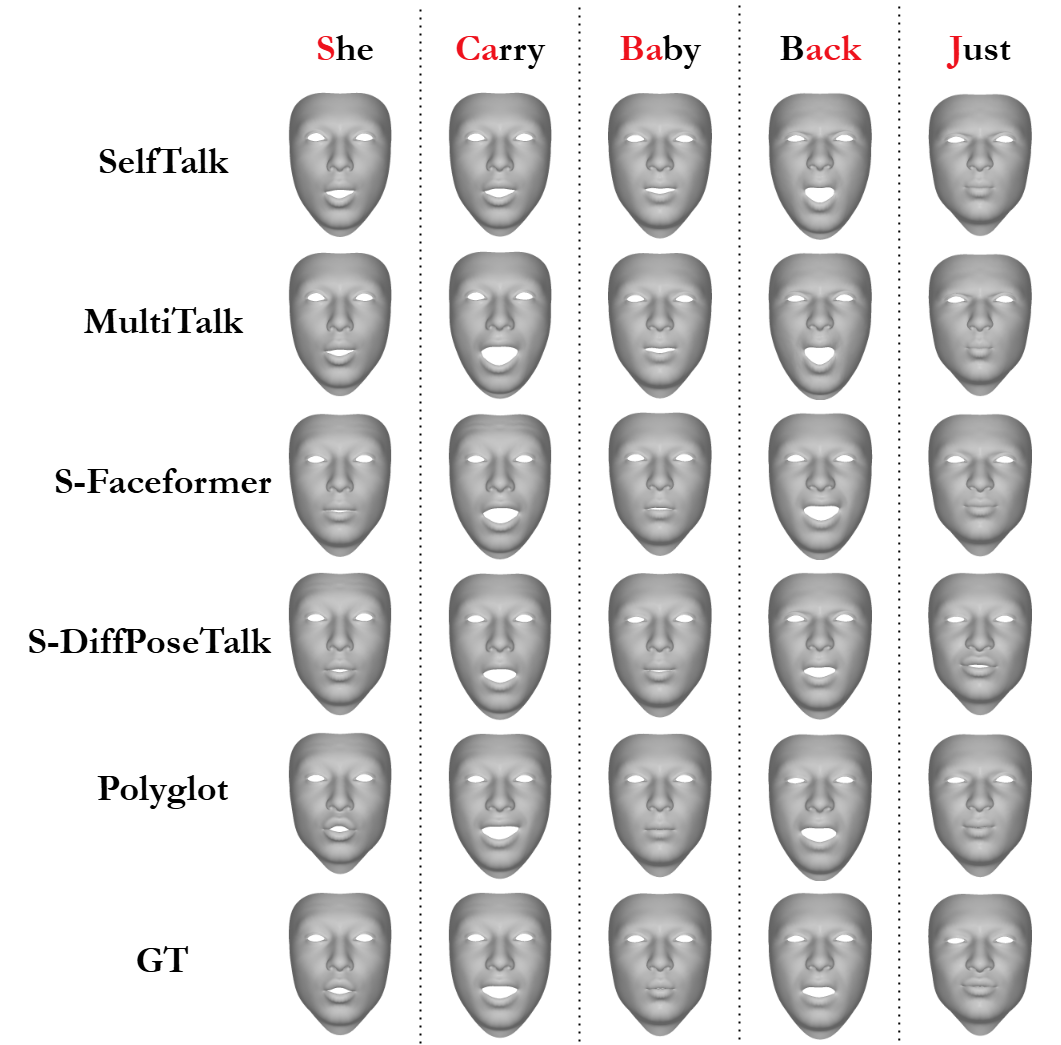}
    \caption{Greyman visualization between predictions and ground truth. We compare Polyglot with state-of-the-art methods}
    \label{fig:comparisons}
    \vspace{-0.5cm}
\end{figure}

In Fig.~\ref{fig:heatmaps}, we present a qualitative comparison of lip synchronization accuracy using heatmaps computed as the \( \ell_2 \) vertex error between predictions and ground truth for a selection of languages. These results contrast our proposed method, Polyglot, with state-of-the-art approaches. We evaluated performance in two languages, Polish and Japanese. In all cases, Polyglot demonstrates better lip synchronization fidelity compared to existing methods. These results further highlight the necessity of language conditioning when extending SDFA models from a single-language to a multilingual setting. Similar findings are illustrated in Fig.~\ref{fig:comparisons}, where we show a greyman comparison between Polyglot and other state-of-the-art methods against the ground-truth from the PolySet dataset. This visualization reinforces the effectiveness of our approach, as Polyglot consistently achieves more accurate lip synchronization than competing methods.
\begin{figure}[ht!]
    \centering
    \includegraphics[width=0.75\linewidth]{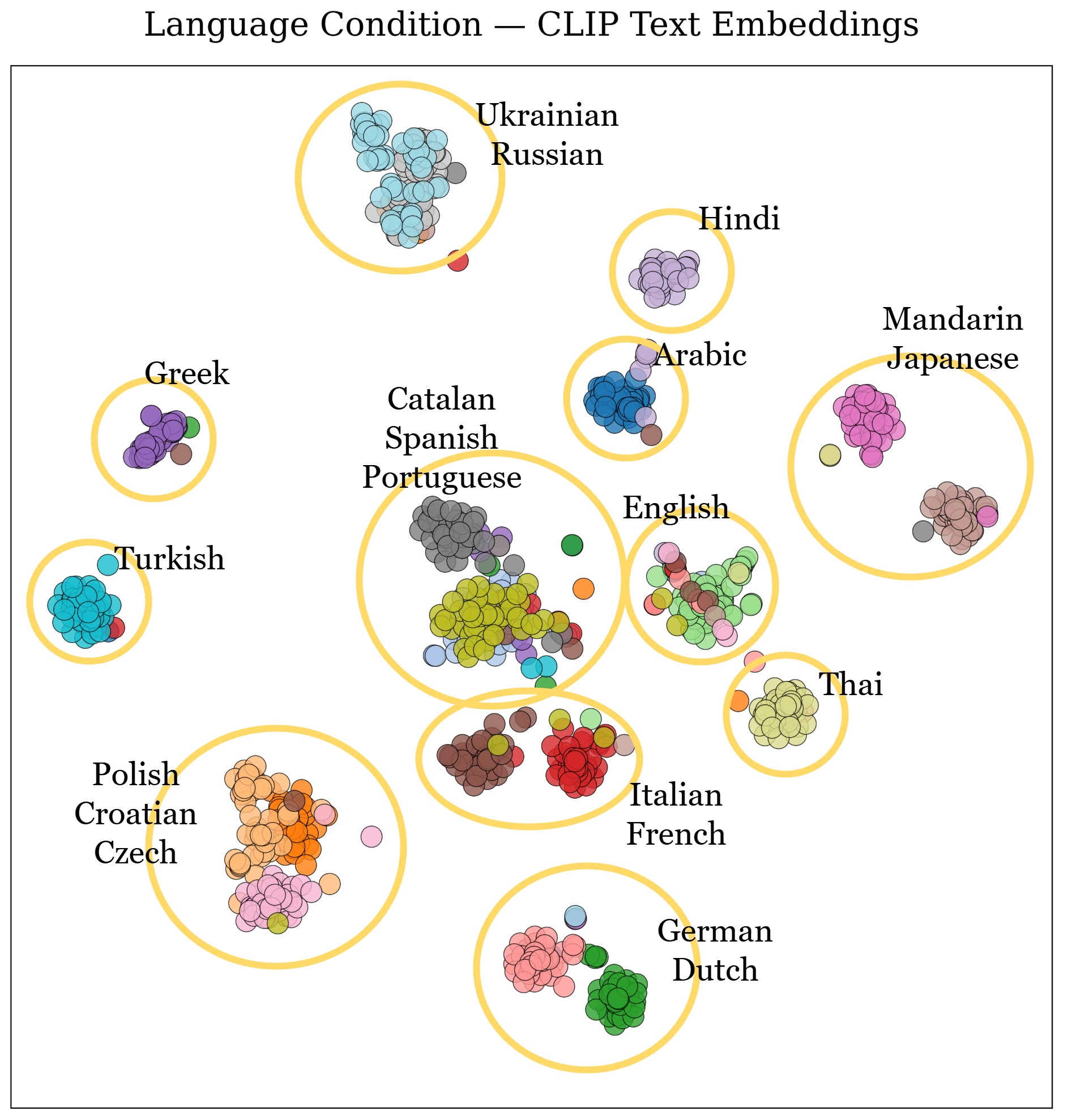}
    \caption{t-SNE visualization of text embeddings produced by concatenating the outputs of the Whisper and CLIP text encoders. 50 audio samples from PolySet for each language are used. 
    }
    \label{fig:tsne}
\end{figure}
\begin{figure}[!ht]
    \centering
    \includegraphics[width=0.95\linewidth]{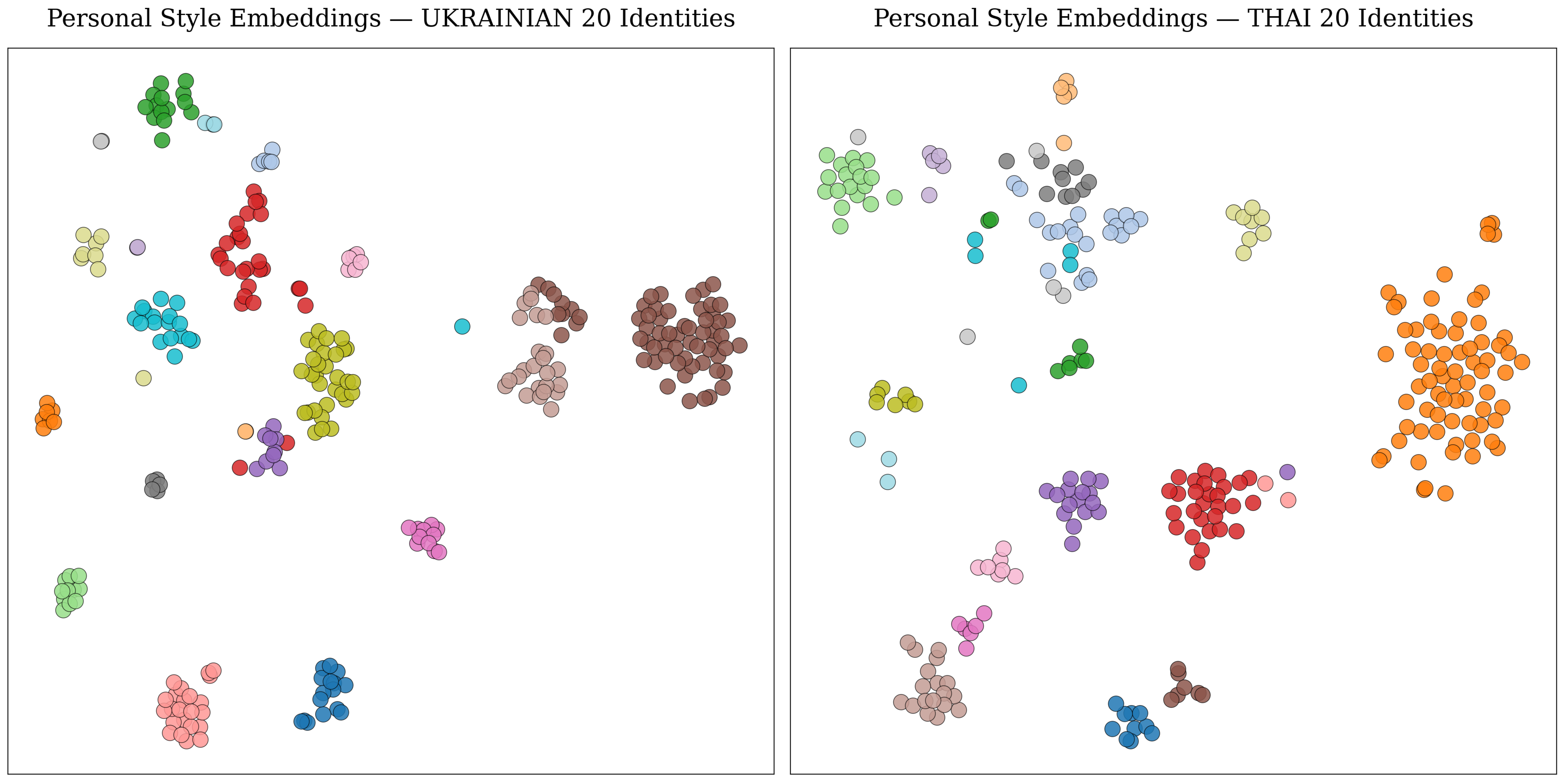}
    \caption{t-SNE plot of personal style embeddings \(S\) for 20 identities across Ukrainian and Thai. The clusters show that the personal speaking style encoder \(E_S\) consistently captures identity-specific personal speaking styles.}
    \label{fig:s_1}
\end{figure}
\begin{figure}[!ht]
    \centering
    \includegraphics[width=0.95\linewidth]{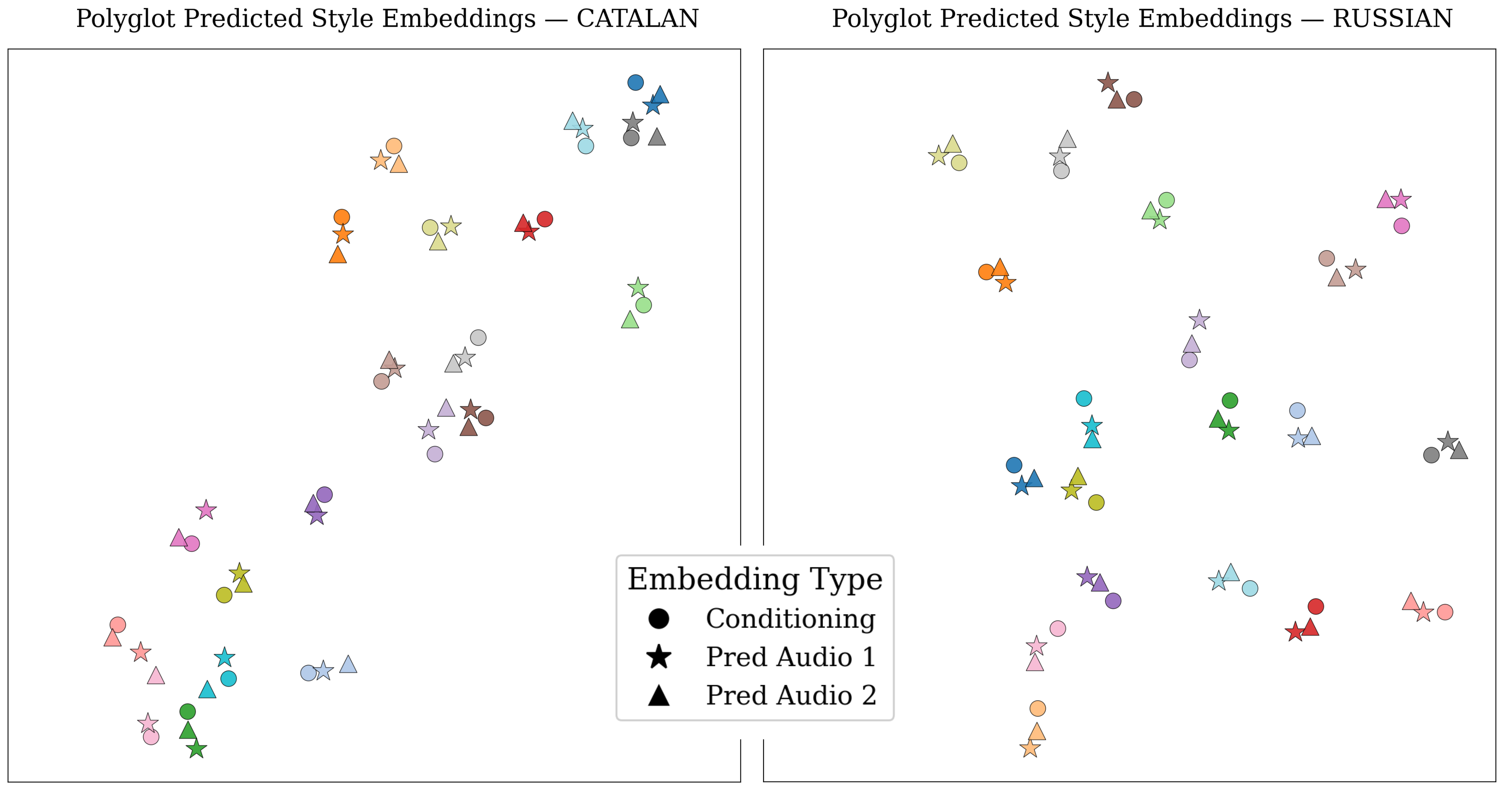}
    \caption{t-SNE visualization of personal style embeddings \(S\) from Polyglot generated sequences with varying audio but fixed personal style input. The overlap with original embeddings shows consistent preservation of personal speaking style.}
    \label{fig:s_2}
\end{figure}

Language styles play a critical rule contributing to the above performance improvement.
As shown in Fig.~\ref{fig:tsne}, latent space of the CLIP text encoder~\cite{clip} used in our method reveals meaningful relationships between languages, with closely related languages such as Polish, Croatian, and Czech forming distinct yet proximate clusters. Similarly, German and Dutch, Mandarin and Japanese, and Italian and French exhibit strong clustering tendencies, while minimal separation is observed between Ukrainian and Russian, as well as Catalan and Spanish. These representations effectively encode linguistic similarities and distinctions, aligning with our objective of capturing cross-linguistic structure.
This behavior can be attributed to multiple factors. First, because transcriptions retain their native scripts~\cite{radford2021learningtransferablevisualmodels}, e.g., Latin, Cyrillic, Arabic, the English-only CLIP model, lacking explicit cross-lingual training, treats script differences as highly discriminative features, leading to well-defined clusters. Conversely, multilingual models are designed to project different languages into a shared semantic space, decreasing the impact of script variations and resulting in more diffuse embeddings.
In addition, tokenization strategies play a critical role in shaping these representations. The English-only CLIP model employs a tokenizer optimized for English, which may tokenize non-Latin scripts suboptimally, leading to higher fragmentation and reinforcing language separability. In contrast, multilingual models leverage subword tokenization techniques that facilitate cross-lingual alignment, effectively smoothing differences across scripts and reducing language-specific clustering. This suggests that while the English-only model capitalizes on surface-level orthographic and phonetic distinctions, multilingual models prioritize deeper semantic alignment. 
To demonstrate our personal style encoder \(E_S\) can effectively extract personal speaking styles from expression sequences, we present a t-SNE visualization in Fig.~\ref{fig:s_1}. The embeddings are derived from 20 different identities, and the visualization shows that the extracted style representations \(S\) form clearly separated clusters, each corresponding to a unique speaker. This confirms that our encoder captures distinctive personal speaking styles.
Furthermore, to validate that Polyglot maintains the intended personal speaking style during generation, we provide another t-SNE visualization in Fig.~\ref{fig:s_2}. This figure compares the conditioning style embeddings with the ones extracted from the model’s predicted expression sequences. In this experiment, the personal speaking style is held constant while the input audio is varied twice using two audio samples in the same language. The results show strong alignment between the conditioning and predicted style embeddings, indicating that the model successfully retains the desired personal speaking style regardless of audio variation. 

\subsection{Ablation Studies}\label{sec:abl}
\begin{table*}[t!]
  \centering
  \caption{Polyglot w/o \(S\) ablations. Comparison of DiffPoseTalk trained in a single-language versus a multilingual setting. We evaluate two versions of Polyglot w/o \(S\): one conditioned on a learnable lookup table of embeddings, similar to MultiTalk, and another conditioned on audio transcript text embeddings. The name in parentheses indicates the PolySet subset used for training each model, while (-) denotes that the single-language DiffPoseTalk model is evaluated only on the language it was trained on.}
    \resizebox{\linewidth}{!}{
    \centering
    \begin{tabular}{@{}l@{\hspace{0.05cm}}ccccccccccccccccccc@{}} 
        \toprule
        & \multicolumn{4}{c}{\textbf{Italian}} &  & \multicolumn{4}{c}{\textbf{Spanish}} &
         & \multicolumn{4}{c}{\textbf{English}} &
         & \multicolumn{4}{c}{\textbf{Italian+Spanish+English}} 
        \\
        \cmidrule{2-5} \cmidrule{7-10} \cmidrule{12-15} \cmidrule{17-20}
         & LVE $\downarrow$ & MVE $\downarrow$ & DTW $\downarrow$ & MOD $\downarrow$ && LVE $\downarrow$ & MVE $\downarrow$ & DTW $\downarrow$ & MOD $\downarrow$ && LVE $\downarrow$ & MVE $\downarrow$ & DTW $\downarrow$ & MOD $\downarrow$ && LVE $\downarrow$ & MVE $\downarrow$ & DTW $\downarrow$ & MOD $\downarrow$\\
        \midrule
        
        DiffPoseTalk (Italian) & 0.280 & 0.058 & \underline{0.162} & \textbf{0.196} && - & - & - & - && - & - & - & - && - & - & - & - \\

        DiffPoseTalk (Spanish) & - & - & - & - && 0.497 & 0.105 & \underline{0.215} & 0.327 && - & - & - & - && - & - & - & - \\

        DiffPoseTalk (English) & - & - & - & - && - & - & - & - && \underline{0.401} & \underline{0.081} & \underline{0.193} & 0.323 && - & - & - & - \\

        DiffPoseTalk (I+E+S) & 0.282 & 0.057 & 0.164 & 0.209 && 0.509 & 0.106 & 0.220 & \underline{0.314} && 0.438 & 0.085 & 0.197 & 0.319 && 0.413 & 0.084 & 0.192 & \underline{0.277} \\

        Polyglot-lt w/o \(S\) (I+E+S) & \underline{0.278} & \underline{0.056} & 0.163 & \underline{0.200} && \underline{0.494} & \underline{0.102} & \underline{0.215} & \textbf{0.313} && 0.417 & 0.082 & 0.194 & \textbf{0.311} && \underline{0.394} & \underline{0.081} & \underline{0.191} & 0.279 \\

        Polyglot w/o \(S\) (I+E+S) & \textbf{0.275} & \textbf{0.055} & \textbf{0.161} & 0.207 && \textbf{0.486} & \textbf{0.101} & \textbf{0.212} & 0.316 && \textbf{0.396} & \textbf{0.080} & \textbf{0.191} & \underline{0.317} && \textbf{0.391} & \textbf{0.080} & \textbf{0.188} & \textbf{0.273} \\
        \bottomrule
        \end{tabular}
        }
        \label{tab:ablations_2}
        \vspace{-0.3cm}
\end{table*}
\begin{table}[ht!]
\vspace{-0.2cm}
  \centering
  \caption{
         Polyglot ablations. We evaluate the effects of personal style conditioning \(S\), language conditioning \(\hat{t}\) , the style loss \(L_{\text{style}}\), and training on fewer languages. Results show each component contributes to improved performance and generalization.
        }
    \resizebox{\linewidth}{!}{
    \centering
    \begin{tabular}{@{}l@{\hspace{0.05cm}}ccccc@{}} 
        \toprule
        & \multicolumn{4}{c}{\textbf{20 Languages}}
        \\
        \cmidrule{2-5} 
         & LVE $\downarrow$ & MVE $\downarrow$ & DTW $\downarrow$ & MOD $\downarrow$ \\
        \midrule
        Polyglot w/o \(S\) & 0.270 & 0.054 & 0.156 & 0.022 \\
        Polyglot w/o \(\hat{t}\) & 0.209 & 0.042 & 0.134 & 0.203 \\
        Polyglot w/o \(L_{style}\) & 0.203 & 0.041 & 0.133 & 0.191\\
        Polyglot (I+E+S+G+F) & 0.303 & 0.061 & 0.162 & 0.247\\
        Polyglot & 0.199 & 0.041 & 0.126 & 0.188 \\
        \bottomrule
        \end{tabular}
        }
        \label{tab:ablations}
\end{table}
We conducted a series of ablation studies to compare different versions of language conditioning for our method. Specifically, we compared Polyglot w/o \(S\) trained on three languages (Italian, Spanish, and English), with DiffPoseTalk, evaluated in both single-language and multilingual settings (Italian, Spanish, and English). As presented in Table ~\ref{tab:ablations_2}, the single-language versions of DiffPoseTalk were assessed on the validation set corresponding to their respective training language. The results in Table~\ref{tab:quantitative} and Table~\ref{tab:ablations_2} show that Polyglot consistently outperforms DiffPoseTalk in both single-language and multilingual training settings, highlighting the value of explicit language conditioning.
Furthermore, Table~\ref{tab:ablations_2} reveals that training a non-multilingual model such as DiffPoseTalk in a multilingual setting results in degraded performance when evaluated on a single-language test set. In contrast, introducing language conditioning into the model improves performance. To validate this, we explored two strategies for incorporating language information: conditioning on text embeddings derived from transcripts and utilizing a learnable lookup table (Polyglot-lt w/o \(S\)) of language embeddings, similar to the approach proposed in MultiTalk~\cite{sungbin2024multitalkenhancing3dtalking}. Results in Table~\ref{tab:quantitative} and Table~\ref{tab:ablations_2} indicate that conditioning on text embeddings leads to superior modeling of language-specific lip movements compared to a lookup table-based approach. 
Furthermore, in Table~\ref{tab:ablations} we present an ablation study on the several architecture components of the model, an ablation study on the \(L_{style}\) loss, and also an ablation to see the model's behavior when training with a lower number of languages. \\
The results show that jointly conditioning on both the transcript-based text embedding \(\hat{t}\) and the personal style embedding \(S\) significantly improves performance, confirming that both components are essential for generating more realistic facial animations.

Furthermore, using the style preservation loss \(L_{\text{style}}\) leads to better outcomes compared to models trained without it. Finally, training the model on a reduced version of Polyset containing only five languages (Italian, English, French, Spanish, and German) results in lower performance than training on the full 20-language Polyset. This underscores the importance of language diversity: increasing the number of training languages enhances generalization and improves overall performance in multilingual SDFA.

\subsection{User Study}
\label{sec:us}
To further evaluate our method, we conducted a user study incorporating human feedback. This evaluation involved $30$ participants and was inspired by prior works~\cite{Fan_Lin_Saito_Wang_Komura_faceformer_2022, facexhubert, peng2023selftalk, richard2021meshtalk, sun2023diffposetalk, nocentini2024scantalk3dtalkingheads}. We designed an A/B test to compare Polyglot against state-of-the-art models, focusing on two key aspects: lip synchronization and the naturalness and realism of facial expressions. Each participant assessed five comparisons between Polyglot and each baseline method. 

\begin{table}[ht!]
\vspace{-0.2cm}
    \centering
    \caption{User study results comparing Polyglot with state-of-the-art models. The values denote the percentages of users who favored Polyglot animations over the competitor.}
    \label{tab:us}
    \resizebox{\linewidth}{!}{
        \begin{tabular}{lc@{\hspace{0.2cm}}c}
        \toprule
        \textbf{Comparison} & \textbf{Lip-sync (\%)} & \textbf{Naturalness (\%)} \\ 
        \midrule
       Polyglot vs. FaceFormer ~\cite{Fan_Lin_Saito_Wang_Komura_faceformer_2022}   & 74.6 & 81.3 \\
       Polyglot vs. SelfTalk ~\cite{peng2023selftalk}  & 60.0 & 60.1 \\
       Polyglot vs. DiffPoseTalk ~\cite{sun2023diffposetalk} & 90.6 & 74.6 \\
       Polyglot vs. MultiTalk ~\cite{sungbin2024multitalkenhancing3dtalking}     & 51.1 & 60.5 \\ 
        \bottomrule
        \end{tabular}
        }
    \vspace{-0.4cm}
\end{table}
Results, presented in Table~\ref{tab:us}, show the percentage of users who preferred Polyglot over the alternatives. 
Results in Table~\ref{tab:us} show that animations generated with Polyglot achieve naturalness and lip-sync quality comparable to state-of-the-art methods. Notably, our approach is preferred over FaceFormer, SelfTalk, and DiffPoseTalk. When compared to MultiTalk, half of the users preferred Polyglot for lip synchronization, while more than half preferred it for naturalness. These results underscore the quality of our generated animations relative to existing speech-driven facial animation methods. In addition, they emphasize the effectiveness of integrating information about language and personal style in a multilingual setting.
\section{Ethical Considerations}
Furthermore, we recognize the ethical implications of generating facial animations. The creation of synthetic narratives using faces carries inherent risks, potentially leading to both intentional and unintentional consequences for individuals and society. We aim for this technology to be used responsibly, ensuring that it promotes ethical and positive applications while mitigating potential misuse.
\section{Conclusions}
\label{sec:conclusions}
We introduce \textbf{Polyglot}, a diffusion-based framework for multilingual speech-driven facial animation that jointly models language and personal speaking style. By combining text embeddings with a style encoder, our method generates realistic, expressive, and temporally coherent animations across diverse languages.
We also present \textbf{Polyset}, a large-scale multilingual dataset covering 20 languages. Extensive experiments show that jointly conditioning on language and style is key to achieving high-fidelity animation, especially in multilingual settings where audio alone is insufficient.
Future work includes improving efficiency and exploring richer style and language representations to further enhance realism and generalization.

\clearpage
\appendix
\section*{Supplementary Material}
\addcontentsline{toc}{section}{Supplementary Material}

\section{3D Face Representation}
In Polyglot, we use a 3D morphable model (3DMM) for 3D face representation. The 3DMM is a statistical deformation model for 3D faces, firstly proposed by Blanz and Vetter~\cite{blanz1999morphable}. Briefly, a 3DMM is built by learning a low-dimensional deformation space from a set of densely registered 3D faces. The learned basis vectors are used to parameterize the shape (and optionally texture) space and synthesize new faces as:
\begin{equation}
    \mathbf{V} = \mathbf{F} + \mathbf{C}\mathbf{a^0},
    \label{eq:3dmm}
\end{equation}

\noindent
where $\mathbf{V}\in \mathbb{R}^{N \times 3}$ is a 3D face of $N$ vertices, $\mathbf{F}\in \mathbb{R}^{N \times 3}$ is a template 3D face, $\mathbf{C}\in \mathbb{R}^{3N \times k}$ are the 3DMM shape bases, and $\mathbf{a^0} \in \mathbb{R}^k$ are the shape coefficients controlling the deformation.\\
Depending on the number of samples, the variability and the type of deformations included in the 3D face training set, different bases can be learned, which define what kind of deformations the learned model will apply. For example, the Basel Face Model~\cite{paysan20093d} can only model global structural facial traits, e.g. thin/large head, as it learned from a set of 200 faces in neutral expression. Other methods such as the DL-3DMM~\cite{ferrari2015dictionary}, the FLAME model~\cite{li2017learning} or ICT~\cite{ict} are instead learned from both neutral and expressive faces. In this way, the 3DMM can also replicate expression and pose related deformations. The expressive power of a 3DMM is also defined by the learning approach used. For example, some explored solutions to learn localized deformation bases~\cite{ferrari2021sparse,neumann2013sparse,luthi2017gaussian}, while others exploited deep models to learn extreme deformations~\cite{bouritsas2019neural}. In this study, we use 3D morphable model ICT with $N = 6706$ vertices whose geometry can be represented using parameters $\{\beta,m,\theta\}$, where $\beta \in \mathbb{R}^{80}$ is the shape (identity) parameter, $m \in \mathbb{R}^{53}$ is the expression parameter, and $\theta \in \mathbb{R}^{6}$ is the head pose parameter. The ICT model allows us to train the proposed method with expression parameters, leading to smaller models and faster inference and training times due to the lower dimensionality of expression parameters. This advantage has also been demonstrated by~\cite{sun2023diffposetalk}, further validating the effectiveness of expression parameters for efficient model training.

\section{Polyglot Implementation Details}
The style autoencoder was trained for 200 epochs on the PolySet dataset using the Adam optimizer with a learning rate of \(10^{-4}\) and a minibatch size of 128.
Polyglot was trained for 1000 epochs on the PolySet dataset using the Adam optimizer with a learning rate of \(10^{-4}\) and a minibatch size of 128. The loss component multipliers were set as follows: \(\lambda_{\text{sim}} = 10\), \(\lambda_{\text{m}} = 10\), \(\lambda_{\text{v}} = 2 \times 10^2\), \(\lambda_{\text{vel}} = 10^2\), and \(\lambda_{\text{s}} = 10^2\). The number of diffusion steps \(N\) was set to 500, with a cosine diffusion noise schedule applied. The model uses \(k = 53\) expression parameters and \(b = 13\) mouth expression parameters. The hidden size \(h\) of the transformer decoder was set to 512, with 6 layers. \(T_w\) and \(T_p\) were set to 75 and 10, respectively. During inference we applied the classifier-free guidance to both audio and text conditioning with a weight of 1.15, as applied in~\cite{sun2023diffposetalk}. Training was performed on a NVIDIA A100-SXM4-80GB GPU.

\section{Additional Quantitative Results}
\label{sec:add_quantitative}
In Table \ref{tab:add_quantitative}, we provide a comprehensive comparison between Polyglot and all state-of-the-art methods across the 15 additional languages in the Polyset dataset that were not included in the main paper.

\begin{table*}[ht!]
    \centering
    \resizebox{\linewidth}{!}{
    \centering
    \begin{tabular}{@{}l@{\hspace{0.1cm}}ccccccccccccccc@{}} 
        \toprule
        & \multicolumn{4}{c}{\textbf{Arabic}} & \phantom{abc} & \multicolumn{4}{c}{\textbf{Catalan}} &
        \phantom{abc} & \multicolumn{4}{c}{\textbf{Croatian}} 
        \\
        \cmidrule{2-5} \cmidrule{7-10} \cmidrule{12-15}
         & LVE $\downarrow$ & MVE $\downarrow$ & DTW $\downarrow$ & MOD $\downarrow$ && LVE $\downarrow$ & MVE $\downarrow$ & DTW $\downarrow$ & MOD $\downarrow$ && LVE $\downarrow$ & MVE $\downarrow$ & DTW $\downarrow$ & MOD $\downarrow$\\
        \midrule
        FaceFormer & 0.329 & 0.064 & 0.170 & 0.260 && 0.280 & 0.054 & 0.162 & 0.217 && 0.239 & 0.046 & 0.150 & 0.185  \\
        
        SelfTalk & 0.494 & 0.097 & 0.202 & 0.265 && 0.352 & 0.070 & 0.176 & 0.205 && 0.335 & 0.068 & 0.171 & 0.203  \\

        DiffPoseTalk & 0.325 & 0.065 & 0.175 & 0.257 && 0.272 & 0.054 & 0.162 & 0.230 && 0.282 & 0.056 & 0.169 & 0.215  \\

        MultiTalk & 0.301 & 0.066 & 0.171 & 0.240 && 0.240 & 0.056 & 0.155 & 0.197 && 0.238 & 0.050 & 0.147 & 0.141  \\
        
        \midrule

        S-Faceformer & 0.282 & 0.531 & 0.156 & 0.233 && 0.252 & 0.048 & 0.157 & 0.218 && 0.226 & 0.044 & 0.148 & 0.167 \\

        S-DiffPoseTalk & \underline{0.253} & \underline{0.050} & \underline{0.150} & \underline{0.207} && \underline{0.239} & \underline{0.047} & \underline{0.142} & \underline{0.184} && \underline{0.194} & \underline{0.040} & \underline{0.137} & \underline{0.153} \\
        
        \midrule

        Polyglot & \textbf{0.203} & \textbf{0.040} & \textbf{0.132} & \textbf{0.193} && \textbf{0.192} & \textbf{0.038} & \textbf{0.127} & \textbf{0.155} && \textbf{0.163} & \textbf{0.033} & \textbf{0.122} & \textbf{0.145} \\
        
        \midrule
        \midrule

        & \multicolumn{4}{c}{\textbf{Czech}} & \phantom{abc} & \multicolumn{4}{c}{\textbf{Dutch}} &
        \phantom{abc} & \multicolumn{4}{c}{\textbf{French}} 
        \\
        \cmidrule{2-5} \cmidrule{7-10} \cmidrule{12-15} 
         & LVE $\downarrow$ & MVE $\downarrow$ & DTW $\downarrow$ & MOD $\downarrow$ && LVE $\downarrow$ & MVE $\downarrow$ & DTW $\downarrow$ & MOD $\downarrow$ && LVE $\downarrow$ & MVE $\downarrow$ & DTW $\downarrow$ & MOD $\downarrow$\\
        \midrule
        FaceFormer & 0.220 & 0.042 & 0.142 & 0.201 && 0.295 & 0.058 & 0.162 & 0.248 && 0.371 & 0.073 & 0.181 & 0.277  \\
        
        SelfTalk  & 0.312 & 0.063 & 0.163 & 0.207 && 0.404 & 0.083 & 0.188 & 0.270 && 0.591 & 0.121 & 0.221 & 0.300  \\

        DiffPoseTalk & 0.242 & 0.048 & 0.150 & 0.206 && 0.324 & 0.066 & 0.177 & 0.240 && 0.334 & 0.068 & 0.175 & 0.248  \\

        MultiTalk & 0.251 & 0.056 & 0.155 & 0.207 && 0.331 & 0.070 & 0.180 & 0.271 && 0.347 & 0.077 & 0.189 & 0.244 \\
        
        \midrule

        S-Faceformer & 0.205 & 0.040 & 0.138 & 0.211 && 0.261 & 0.051 & 0.154 & 0.248 && 0.339 & 0.068 & 0.172 & 0.276 \\

        S-DiffPoseTalk & \underline{0.179} & \underline{0.036} & \underline{0.129} & \underline{0.168} && \underline{0.239} & \underline{0.049} & \underline{0.146} & \underline{0.200} && \underline{0.322} & \underline{0.066} & \underline{0.162} & \underline{0.230} \\
        
        \midrule

        Polyglot & \textbf{0.146} & \textbf{0.029} & \textbf{0.112} & \textbf{0.157} && \textbf{0.215} & \textbf{0.044} & \textbf{0.135} & \textbf{0.191} && \textbf{0.254} & \textbf{0.052} & \textbf{0.140} & \textbf{0.205} \\
        
        \midrule
        \midrule

        & \multicolumn{4}{c}{\textbf{German}} & \phantom{abc} & \multicolumn{4}{c}{\textbf{Greek}} &
        \phantom{abc} & \multicolumn{4}{c}{\textbf{Hindi}} 
        \\
        \cmidrule{2-5} \cmidrule{7-10} \cmidrule{12-15} 
         & LVE $\downarrow$ & MVE $\downarrow$ & DTW $\downarrow$ & MOD $\downarrow$ && LVE $\downarrow$ & MVE $\downarrow$ & DTW $\downarrow$ & MOD $\downarrow$ && LVE $\downarrow$ & MVE $\downarrow$ & DTW $\downarrow$ & MOD $\downarrow$\\
        \midrule
        FaceFormer & 0.295 & 0.056 & 0.159 & 0.238 && 0.264 & 0.054 & 0.143 & 0.271 && 0.230 & 0.045 & 0.144 & 0.225  \\
        
        SelfTalk  & 0.412 & 0.083 & 0.188 & 0.244 && 0.327 & 0.069 & 0.164 & 0.303 && 0.269 & 0.056 & 0.156 & 0.221  \\

        DiffPoseTalk & 0.354 & 0.070 & 0.183 & 0.266 && 0.280 & 0.058 & 0.155 & 0.303 && 0.233 & 0.047 & 0.149 & 0.226  \\

        MultiTalk & 0.287 & \underline{0.051} & 0.155 & 0.255 && 0.247 & 0.059 & 0.140 & 0.251 && 0.220 & 0.049 & 0.144 & 0.220 \\
        
        \midrule

        S-Faceformer & 0.248 & 0.050 & 0.153 & 0.232 && 0.233 & \underline{0.047} & \underline{0.134} & 0.263 && 0.220 & 0.043 & 0.146 & 0.227 \\

        S-DiffPoseTalk & \underline{0.259} & 0.053 & \underline{0.148} & \underline{0.202} && \underline{0.221} & \underline{0.047} & 0.135 & \underline{0.225} && \underline{0.198} & \underline{0.037} & \underline{0.129} & \underline{0.187} \\
        
        \midrule

        Polyglot & \textbf{0.239} & \textbf{0.049} & \textbf{0.139} & \textbf{0.191} && \textbf{0.183} & \textbf{0.038} & \textbf{0.115} & \textbf{0.214} && \textbf{0.142} & \textbf{0.029} & \textbf{0.109} & \textbf{0.165} \\
        
        \midrule
        \midrule

        & \multicolumn{4}{c}{\textbf{Mandarin}} & \phantom{abc} & \multicolumn{4}{c}{\textbf{Polish}} &
        \phantom{abc} & \multicolumn{4}{c}{\textbf{Russian}} 
        \\
        \cmidrule{2-5} \cmidrule{7-10} \cmidrule{12-15} 
         & LVE $\downarrow$ & MVE $\downarrow$ & DTW $\downarrow$ & MOD $\downarrow$ && LVE $\downarrow$ & MVE $\downarrow$ & DTW $\downarrow$ & MOD $\downarrow$ && LVE $\downarrow$ & MVE $\downarrow$ & DTW $\downarrow$ & MOD $\downarrow$\\
        \midrule
        FaceFormer & 0.256 & 0.049 & 0.148 & 0.242 && 0.253 & 0.049 & 0.148 & 0.225 && 0.206 & 0.038 & 0.134 & 0.202  \\
        
        SelfTalk  & 0.354 & 0.073 & 0.173 & 0.260 && 0.351 & 0.075 & 0.167 & 0.258 && 0.298 & 0.060 & 0.154 & 0.215  \\

        DiffPoseTalk & 0.224 & 0.046 & 0.147 & 0.221 && 0.251 & 0.052 & 0.152 & 0.229 && 0.226 & 0.046 & 0.147 & 0.206  \\

        MultiTalk & 0.223 & 0.052 & 0.151 & 0.214 && 0.239 & 0.047 & 0.139 & 0.218 && 0.210 & 0.045 & 0.138 & 0.171 \\
        
        \midrule

        S-Faceformer & 0.199 & 0.040 & 0.136 & 0.227 && \underline{0.191} & \underline{0.038} & \underline{0.131} & 0.205 && 0.162 & 0.038 & 0.131 & 0.205 \\

        S-DiffPoseTalk & \underline{0.186} & \underline{0.038} & \underline{0.125} & \underline{0.191} && 0.199 & 0.041 & \underline{0.131} & \underline{0.201} && \underline{0.156} & \underline{0.032} & \underline{0.117} & \underline{0.158} \\
        
        \midrule

        Polyglot & \textbf{0.153} & \textbf{0.031} & \textbf{0.111} & \textbf{0.172} && \textbf{0.169} & \textbf{0.035} & \textbf{0.115} & \textbf{0.176} && \textbf{0.133} & \textbf{0.027} & \textbf{0.104} & \textbf{0.152} \\
        
        \midrule
        \midrule

        & \multicolumn{4}{c}{\textbf{Thai}} & \phantom{abc} & \multicolumn{4}{c}{\textbf{Turkish}} &
        \phantom{abc} & \multicolumn{4}{c}{\textbf{Ukrainian}} 
        \\
        \cmidrule{2-5} \cmidrule{7-10} \cmidrule{12-15} 
         & LVE $\downarrow$ & MVE $\downarrow$ & DTW $\downarrow$ & MOD $\downarrow$ && LVE $\downarrow$ & MVE $\downarrow$ & DTW $\downarrow$ & MOD $\downarrow$ && LVE $\downarrow$ & MVE $\downarrow$ & DTW $\downarrow$ & MOD $\downarrow$\\
        \midrule
        FaceFormer & 0.298 & 0.059 & 0.165 & 0.258 && 0.327 & 0.064 & 0.168 & 0.246 && 0.179 & 0.035 & 0.125 & 0.190  \\

        SelfTalk  & 0.398 & 0.079 & 0.187 & 0.276 && 0.531 & 0.108 & 0.209 & 0.289 && 0.180 & 0.038 & 0.128 & 0.185  \\
        
        DiffPoseTalk & 0.292 & 0.059 & 0.167 & 0.257 && 0.384 & 0.076 & 0.185 & 0.252 && 0.175 & 0.036 & 0.129 & 0.190  \\
        
        MultiTalk & 0.271 & 0.055 & 0.169 & 0.241 && 0.322 & 0.069 & 0.167 & 0.242 && 0.170 & 0.033 & 0.127 & 0.171 \\
        
        \midrule

        S-Faceformer & 0.254 & 0.052 & 0.154 & 0.236 && 0.299 & 0.059 & 0.161 & 0.222 && 0.160 & 0.031 & 0.123 & 0.178 \\

        S-DiffPoseTalk & \underline{0.253} & \underline{0.051} & \underline{0.152} & \underline{0.228} && \underline{0.258} & \underline{0.052} & \underline{0.148} & \underline{0.210} && \underline{0.146} & \underline{0.029} & \underline{0.111} & \underline{0.167} \\
        
        \midrule

        Polyglot & \textbf{0.218} & \textbf{0.045} & \textbf{0.134} & \textbf{0.202} && \textbf{0.242} & \textbf{0.049} & \textbf{0.137} & \textbf{0.197} && \textbf{0.114} & \textbf{0.023} & \textbf{0.094} & \textbf{0.152} \\
        
        \bottomrule
        \end{tabular}
        }
        \caption{Polyglot in comparison with SOTA methods, all models listed in the table were trained on a multilingual dataset spanning 20 languages. We then evaluated each model in both monolingual and multilingual settings to assess their performance across different linguistic conditions.}
        \label{tab:add_quantitative}
\end{table*}

{\small
\bibliographystyle{ieee}
\bibliography{egbib}
}

\end{document}